\pdfoutput=1
\documentclass[letterpaper]{article} 
\usepackage{aaai24}  
\usepackage{times}  
\usepackage{helvet}  
\usepackage{courier}  
\usepackage[hyphens]{url}  
\usepackage{graphicx} 
\usepackage{natbib}  
\usepackage{caption} 
\usepackage{amsmath}
\usepackage{amssymb}
\usepackage{mathtools}
\usepackage{booktabs}
\usepackage{tabularx}
\usepackage[export]{adjustbox}
\usepackage{multirow}
\usepackage{colortbl}
\usepackage{xcolor} 
\usepackage{bbding}
\usepackage{amssymb}
\usepackage{pifont}
\usepackage{tikz}

\definecolor{mygray}{gray}{.9}
\definecolor{myyellow}{rgb}{0.71, 0.55, 0.0}
\definecolor{myblue}{rgb}{0.0, 0.71, 0.71}
\definecolor{color2}{rgb}{0.55, 0.71, 0.0}
\definecolor{color1}{rgb}{0.98, 0.81, 0.69}
\definecolor{color3}{rgb}{1.0, 0.6, 0.4}
\definecolor{color4}{rgb}{0.29, 0.59, 0.82}

\newcommand{\encoder}{\mathcal{E}}
\newcommand{\decoder}{\mathcal{D}}
\newcommand{\noiser}{\mathcal{A}}

\newcommand{\hpixel}{H}
\newcommand{\wpixel}{W}
\newcommand{\cpixel}{3}

\newcommand{\expec}{\mathbb{E}}

\newcommand{\model}{\epsilon_\theta}
\newcommand{\conditioner}{\tau_\theta}

\newcommand{\cond}{y}
\newcommand{\zt}[1]{z_{#1}}

\makeatletter
\newif\if@restonecol
\makeatother

\usepackage[linesnumbered,ruled,vlined]{algorithm2e}
\usepackage{algpseudocode}
\usepackage{amsmath}

\usepackage{newfloat}
\usepackage{listings}
\DeclareCaptionStyle{ruled}{labelfont=normalfont,labelsep=colon,strut=off} 
\title{Any-Size-Diffusion: Toward Efficient Text-Driven \\ Synthesis for Any-Size HD Images}
\author{
    Qingping Zheng\textsuperscript{\rm 1, 2 *},
    Yuanfan Guo\textsuperscript{\rm 2 *},
    Jiankang Deng\textsuperscript{\rm 3}, \\
    Jianhua Han\textsuperscript{\rm 2},
    Ying Li\textsuperscript{\rm 1 \dag},
    Songcen Xu\textsuperscript{\rm 2},
    Hang Xu\textsuperscript{\rm 2 \dag}
}
\affiliations{
    \textsuperscript{\rm 1}Northwestern Polytechnical University\\
    \textsuperscript{\rm 2}Huawei Noah’s Ark Lab\\
    \textsuperscript{\rm 3}Huawei UKRD\\
}

\usepackage{bibentry}

\begin{document}

\maketitle


\begin{abstract}
Stable diffusion, a generative model used in text-to-image synthesis, frequently encounters resolution-induced composition problems when generating images of varying sizes.
This issue primarily stems from the model being trained on pairs of single-scale images and their corresponding text descriptions. 
Moreover, direct training on images of unlimited sizes is unfeasible, as it would require an immense number of text-image pairs and entail substantial computational expenses.
To overcome these challenges, we propose a two-stage pipeline named \textit{\textbf{A}ny-\textbf{S}ize-\textbf{D}iffusion} (\textbf{ASD}), designed to efficiently generate well-composed images of any size, while minimizing the need for high-memory GPU resources.
Specifically, the initial stage, dubbed Any Ratio Adaptability Diffusion (ARAD), leverages a selected set of images with a restricted range of ratios to optimize the text-conditional diffusion model, thereby improving its ability to adjust composition to accommodate diverse image sizes.
To support the creation of images at any desired size, we further introduce a technique called Fast Seamless Tiled Diffusion (FSTD) at the subsequent stage. This method allows for the rapid enlargement of the ASD output to any high-resolution size, avoiding seaming artifacts or memory overloads.
Experimental results on the LAION-COCO and MM-CelebA-HQ benchmarks demonstrate that ASD can produce well-structured images of arbitrary sizes, cutting down the inference time by $2\times$ compared to the traditional tiled algorithm.

\end{abstract}

\begin{figure}[h]
    \centering
    \includegraphics[width=1.0\linewidth]{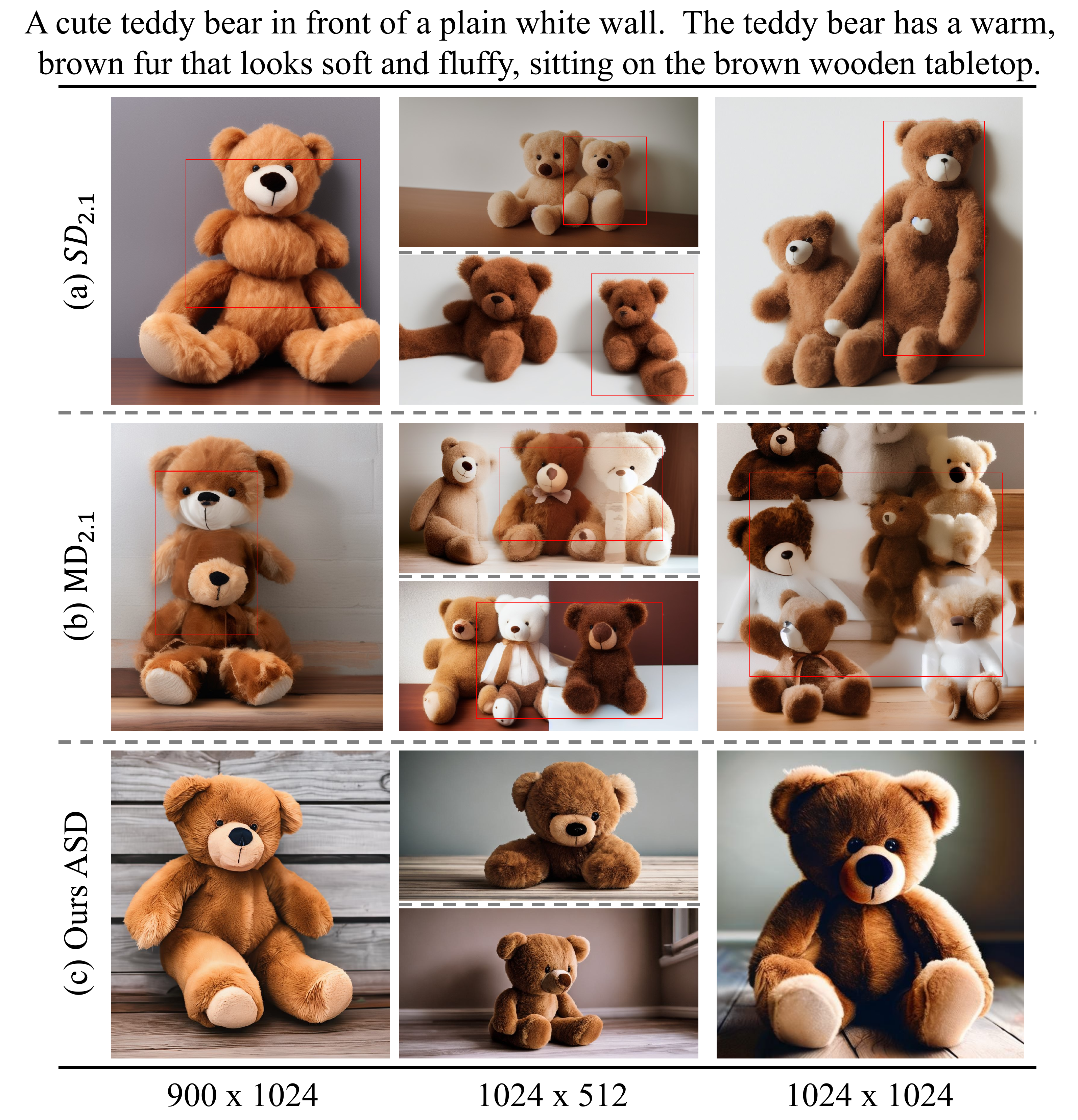}
    \vspace{-6mm}
    \caption{\textbf{Resolution-induced Poor Composition.} Given the text, (a) SD$_{2.1}$ and (b) MD$_{2.1}$, a MultiDiffusion model, raise poor composition issues in {\color{red}{red boxes}} when synthesizing images of varying sizes, as opposed to (c) our ASD.
    \vspace{-8.0mm}}
    \label{problem}
\end{figure}

\vspace{-4.0mm}
\section{Introduction}

In text-to-image synthesis, Stable Diffusion (SD) \cite{Rombach_2022_CVPR} has emerged as a significant advancement. 
Existing SD models \cite{Ruiz_2023_CVPR,Meng_2023_CVPR} transform text aligned with image components into high-quality images, typically sized at $512 \times 512$ pixels. 
Despite these models having the ability to handle varying sizes, they noticeably struggle with resolution changes, resulting in poor composition (\textit{e.g.}, improper cropping and unnatural appearance), a problem demonstrated in Figure~\ref{problem}(a). 
The root of this issue lies in the models trained mainly on pairs of text and images of a uniform size, overlooking the complexities of handling images at multiple resolutions. Consequently, this leads to observed deficiencies in image composition.

In pursuit of generating well-structured images at arbitrary aspect ratios, guided by textual descriptions, the MultiDiffusion methodology \cite{bar2023multidiffusion} leverages a pretrained text-conditional diffusion  (\textit{e.g.}, stable diffusion), as a reference model and controls image synthesis through the utilization of several reference diffusion processes.
Remarkably, the entire process is realized without requiring further training or fine-tuning. 
While efficient, it does not completely resolve the limitations associated with handling the reference model's multi-resolution images. As a result, the production of images may exhibit suboptimal compositional quality.
The underlying reason is also tied to the reference model's training on images constrained to a single-scale size, as illustrated in Figure~\ref{problem}(b).


A direct and appealing solution to the problem is to train the SD model to cope with every possible image size. Yet, this approach encounters an immediate and significant barrier: the infinite diversity of image ratios, which makes it practically unfeasible.
Furthermore, it's challenging to gather an extensive collection of high-resolution images and corresponding text pairs.
Even with a plethora of high-quality datasets available, the intrinsic pixel-based nature of SD requires substantial computational resources, particularly when dealing with high-resolution images of various sizes.
The problem is further aggravated when considering the use of megapixel images for SD training, as this involves extensive repeated function equations and gradient computations in the high-dimensional space of RGB images \cite{NIPS2020_diffusion}.
Even when a trained model is ready, the inference step is also time-consuming and memory-intensive.
Through empirical observation, we have found that attempts to generate 4K HD images using the SD model trigger out-of-memory errors when executed on a GPU with a 32GB capacity.

The key insight of this paper is to introduce a pioneering \emph{\textbf{A}ny-\textbf{S}ize-\textbf{D}iffusion} (\textbf{ASD}) model, executed in two stages, which has the capability to synthesize high-resolution images of arbitrary sizes from text prompts. 
In its dual-phase approach, our ASD not only efficiently handles the resolution-induced poor composition but also successfully circumvents out-of-memory challenges.
At the outset, we are faced with the complexity of accommodating all conceivable image sizes, a challenge that might seem intractable. To address this, in the first stage, we introduce a multi-aspect ratio training strategy that operates within a well-defined, manageable range of ratios. This strategy is used to optimize our proposed \textit{Any Ratio Adaptability Diffusion} (ARAD) model. As a result, it enables the production of well-composed images that are adaptable to any size within a specified range, while also ensuring a reduced consumption of GPU memory.
In order to yield images that can fit any size, in the second stage, we propose an additional method called \textit{Fast Seamless Tiled Diffusion} (FSTD) to magnify the image output originating from the preceding ARAD.
Contrary to the existing tiled diffusion methods~\cite{jiménez2023mixture}, which address the seaming issue but compromise on the speed of inference, our proposed FSTD designs an implicit overlap within the tiled sampling diffusion process. This innovation manages to boost inference speed without the typical seaming problems, achieving the high-fidelity image magnification.
To sum up, the contributions of this paper are as follows:
\begin{itemize}
    \item We are the first to develop the \emph{\textbf{A}ny-\textbf{S}ize-\textbf{D}iffusion} (\textbf{ASD}) model, a two-phase pipeline that synthesizes high-resolution images of any size from text, addressing both composition and memory challenges.
    
    \item We introduce a multi-aspect ratio training strategy, implemented within a defined range of ratios, to optimize ARAD, allowing it to generate well-composed images adaptable to any size within a specified range.
    
    \item We propose an implicit overlap in FSTD to enlarge images to arbitrary sizes, effectively mitigating the seaming problem and simultaneously accelerating the inference time by 2$\times$ compared to the traditional tiled algorithm. 
\end{itemize}


\section{Related Work}

\paragraph{Stable Diffusion.} 
Building upon the foundations laid by the Latent Diffusion Model (LDM) \cite{Rombach_2022_CVPR}, diffusion models \cite{NIPS2020_diffusion,song2021scorebased} have achieved substantial success across various domains,
including text-to-image generation \cite{nichol2022glide,ramesh2022hierarchical,saharia2022photorealistic}, image-to-image translation \cite{dhariwal2021diffusion,nichol2021improved}, and multi-modal generation \cite{ruan2023mm}.
Owing to their robust ability to capture complex distributions and create diverse, high-quality samples, diffusion models excel over other generative methods \cite{NIPS2014_5ca3e9b1}.
In the field, Stable Diffusion (SD) \cite{Rombach_2022_CVPR} has emerged as a leading model for generating photo-realistic images from text. While adept at producing naturalistic images at certain dimensions (\textit{e.g.}, $512 \times 512$), it often yields unnatural outputs with sizes beyond this threshold.
This constraint principally originates from the fact that existing stable diffusion models are exclusively trained on images of a fixed size, leading to a deficiency in high-quality composition on other sizes.
In this paper, we introduce our Any-Size-Diffusion (ASD) model, designed to generate high-fidelity images without size constraints.\\


\paragraph{Diffusion-based Image Super-Resolution.} 
The objective of Image Super-Resolution (SR) is 
infer a high-resolution (HR) image from a corresponding low-resolution (LR) counterpart.
The utilization of generative models to magnify images often omits specific assumptions about degradation, leading to challenging situations in real-world applications.
Recently, diffusion-based methods \cite{sahak2023denoising,saharia_pami_2022,li2023dissecting,ma2023solving} have shown notable success in real-world SR by exploiting generative priors within these models.
Though effective, these approaches introduce considerable computational complexity during training, with a quadratic increase in computational demands as the latent space size increases.
An optimized method, known as StableSR\cite{wang2023exploiting}, was developed to enhance performance while reducing GPU memory consumption. However, this method can become time-inefficient when processing images divided into numerous overlapping regions.
In the next phase of our ASD pipeline, we present a fast seamless tiled diffusion technique, aimed at accelerating inference time in image SR. 

\begin{figure*}[h]
    \centering
    \includegraphics[width=1.0\linewidth]{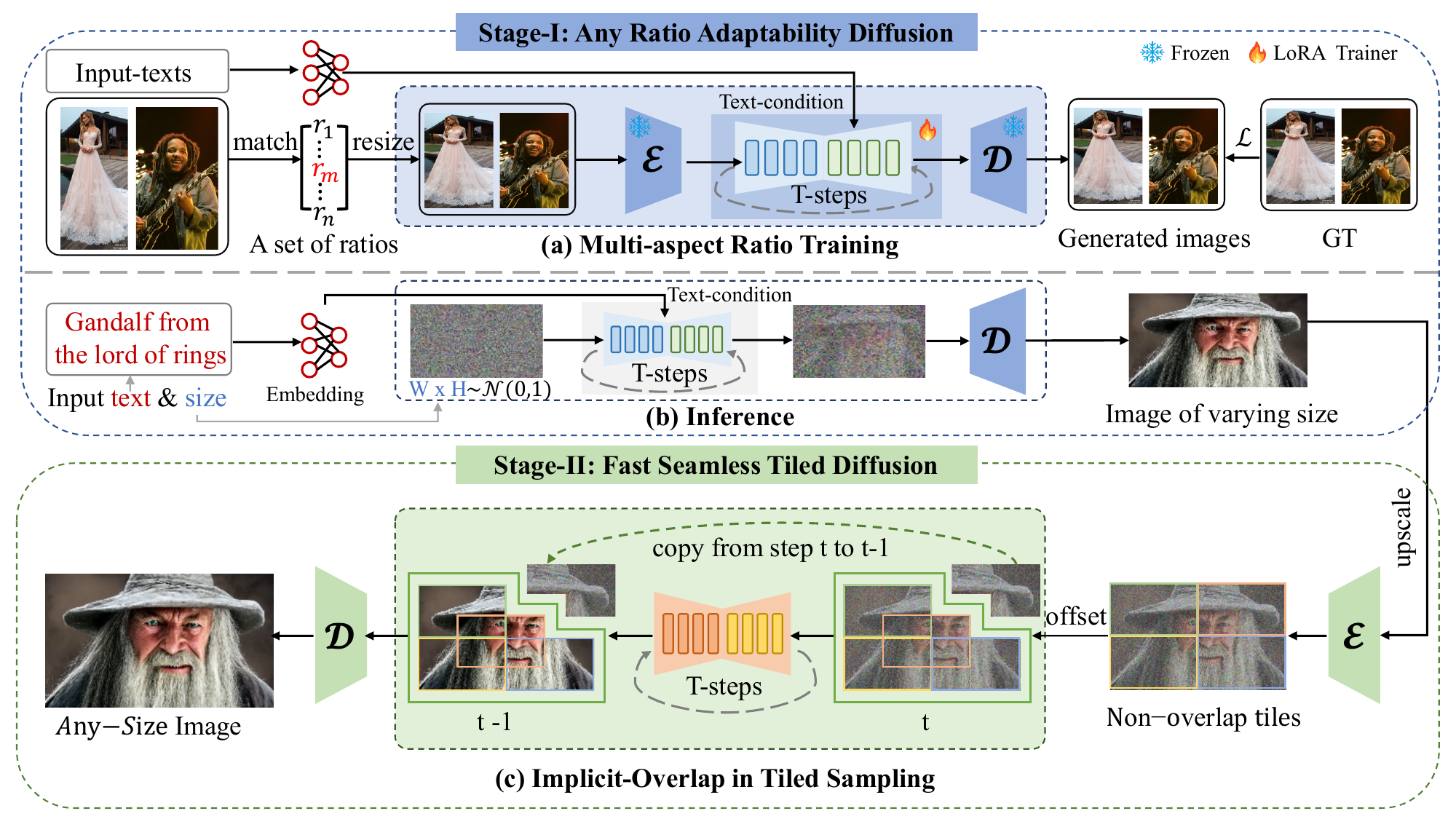}
    \vspace{-6.0mm}
    \caption{\textbf{The Any-Size-Diffusion (ASD) pipeline}, including: 1) Stage-I, Any Ratio Adaptability Diffusion, translates text into images, adapting to various aspect ratios, and 2) is responsible for transforming low-resolution images from the Stage-I into high-resolution versions of any specified size. For procedure (c), the implicit overlap in tiled sampling, only the solid green line region is sent to the UNetModel for current denoising. At Stage-II, the dashed green arrow represents regions that are directly copied from the preceding denoised latents, potentially enhancing efficiency and consistency within the overall process.}
    \label{framework}
    \vspace{-1.0em}
\end{figure*}

\section{Method}

To resolve the issue of resolution-induced poor composition when creating high-fidelity images of various sizes from any text prompt, we propose a straightforward yet efficient approach called \emph{Any Size Diffusion} (ASD).
This approach simplifies the process of text-to-image synthesis by breaking it down into two stages (see Figure~\ref{framework}).
\begin{itemize}
    \item \textbf{Stage-I, termed as Any Ratio Adaptability Diffusion (ARAD)}, trains on multiple aspect-ratio images and generates an image conditioned on a textual description and noise size, avoiding poor composition issues.
    \item \textbf{Stage-II, known as Fast Seamless Tiled Diffusion (FSTD)}, magnifies the image from Stage-I to a predetermined larger size, ultimately producing a high-resolution synthesized image, adjustable to any specified size.
\end{itemize}

\subsection{Pipeline}


As depicted in Figure~\ref{framework}, ARAD is implemented based on the text-conditioned latent diffusion model (LDM) \cite{Rombach_2022_CVPR} for arbitrary ratio image synthesis.
During the inference process, ARAD receives a user-defined text prompt and noise size (\textit{e.g., }``Gandalf from the lord for the rings"). Initially, the pre-trained text encoder \cite{Cherti_2023_CVPR} is adapted to process this prompt, subsequently generating a contextual representation referred to as a textual embedding $\tau_\theta(y)$.
Then, a random noise of the base resolution size, denoted as $\epsilon$, is initialized.
The noisy input conditioned on the textual embedding $p(\epsilon|y)$ is progressively denoised by the UNetModel \cite{Cherti_2023_CVPR}. This process is iterated through $T$ times, leveraging the DDPM algorithm \cite{song2020denoising} to continuously remove noises and restore the latent representation $z$. Ultimately, a decoder $\decoder$ is employed to convert the denoised latent back into an image $I \in \mathbb{R}^{H \times W \times 3}$. Here, $H$ and $W$ represent the height and width of the image, respectively.


Subsequently, the FSTD model accepts the image generated in the previous step as input and performs inference based on the image-conditional diffusion \cite{wang2023exploiting}.  In detail, the image is magnified by a specified size. A pretrained visual encoder $\encoder$ is employed to map the resulting image $I' \in \mathbb{R}^{H' \times W' \times 3}$ into a latent representation $z =\encoder(I')$. A normal distribution-based noise $\epsilon \sim \mathcal{N}(0, 1)$ is then added to it, yielding the noisy latent variable $z'=\noiser(z)$.
The image, conditioned on itself $p(z'|z)$, undergoes progressive iterations by the UNetModel, utilizing our proposed tiled sampling $I \in \mathbb{R}^{H \times W \times 3}$ for $T$ cycles.
Lastly, the decoder $\decoder$ is employed to project the denoised latent variable into the final output, effectively transforming the latent space back into the image domain.
\subsection{Any Ratio Adaptability Diffusion (ARAD)}
In this stage, ARAD is proposed to make the model have the capability of generating an image, adjustable to varying aspect ratios, resolving the issue of resolution-induced poor composition. 
This stage is mainly achieved by our designed multi-aspect ratio training strategy.



\noindent\textbf{Multi-aspect ratio training.}
Instead of directly training on the original image and text pairs, we employ our aspect-ratio strategy to map the original image into an image with a specific ratio.
To be more precise, we define a set of ratios $\{r_1, r_2, ..., r_n\}$, each corresponding to specific sizes $\{s_1, s_2, ..., s_n\}$, where $n$ represents the number of predefined aspect ratios.
For each training image $x \in \mathbb{R}^{\hpixel \times \wpixel \times \cpixel}$, we calculate the image ratio as $r = H/W$. This ratio $r$ is then compared with each predefined ratio, selecting the one $m$ with the smallest distance as the reference ratio.
The index $m$ is determined by
\begin{equation}
\arg \min_{m} f(m) = \{|r_1 - r|, \cdots, |r_m - r|, \cdots, |r_n - r|\},
\end{equation}
where $f(m)$ represents the smallest distance between the current ratio and the predefined ratio. Therefore, if the image has a ratio similar to the $m^{th}$ predefined size $s_m$, the original size of the training image is resized to $s_m$.

\noindent\textbf{Forward ARAD process.} During the training process,
a pair consisting of an image and its corresponding text $(x, y)$ is processed, where $x$ represents an image in the RGB space $\mathbb{R}^{\hpixel \times \wpixel \times \cpixel}$, and $y$ denotes the associated text.
A fixed visual encoder, $\encoder$, is used to transform the resized image $s_m$ into a spatial latent code $z$. Meanwhile, the corresponding text is converted into a textual representation $\tau_\theta(y)$ via OpenCLIP \cite{Cherti_2023_CVPR}. 
For the total steps $T$, conditional distributions of the form $p(z_{t} \vert \cond)$, $t = 1 \cdots T$, can be modeled using a denoising autoencoder $\model(\zt{t},t,\cond)$. Consequently, the proposed ARAD can be learned using an objective function
\begin{equation}
L_{ARAD} = \expec_{\encoder(x), y, \epsilon \sim \mathcal{N}(0, 1), t }\Big[ \Vert \epsilon - \model(z_{t},t, \conditioner(y)) \Vert_{2}^{2}\Big] \, .
\label{eq:cond_loss}
\end{equation}




\subsection{Fast Seamless Tiled Diffusion (FSTD)}

In the second stage, we propose FSTD, a training-free approach built on StableSR \cite{wang2023exploiting} that amplifies the ARAD-generated image to any preferred size.
To achieve efficient image super-resolution without heavy computational demands during inference, 
we devise an implicit overlap technique within the tiled sampling method.

\begin{figure}
    \centering
    \includegraphics[width=1.0\linewidth]{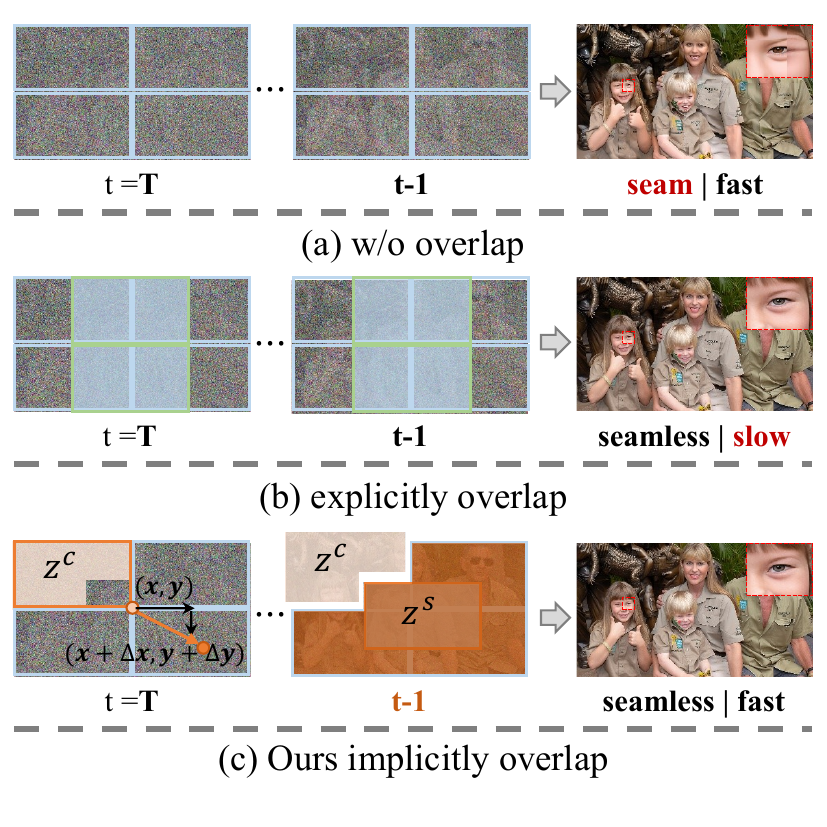}
    \vspace{-12.0mm}
    \caption{\textbf{Comparison of various tiling strategies}: (a) without overlapping, (b) with explicit overlapping, and (c) with implicit overlapping. Green tiles are explicit overlaps, and the orange tile is our implicit overlap at step t-1.
    }
    \label{tiled_diffusion}
    \vspace{-8.0mm}
\end{figure}

\noindent\textbf{Tiled sampling.}
For clarity, consider an upscaled image $\mathcal{I} \in \mathbb{R}^{H' \times W' \times \cpixel}$, partitioned into $M$ petite tiles, symbolized as $\{P_i \in \mathbb{R}^{h \times w \times \cpixel} \ | \ 1\leq i \leq M\}$, where $w$ and $h$ denote the width and height of each tile. We initially encode each tile $P_i$ using an encoder function $\encoder$, adding the random noise, to generate a set of noisy latent representations $\mathcal{Z} = \{\mathcal{Z}_i = \encoder(P_i) + \epsilon_i \ | \ \epsilon_i \sim \mathcal{N}(0, 1), \ 1 \leq i \leq M \}$. Subsequently, each noisy tile is processed by the UNetModel conditioned on the original tile for $T$ steps, resulting in a set of denoised latents $\mathcal{Z'} = \{\mathcal{Z'}_i| \ \epsilon_i \sim \mathcal{N}(0, 1), \ 1 \leq i \leq M \}$. Finally, a decoder $f_\decoder$ is applied to convert them back into image space, culminating in the reconstructed image
\begin{equation}
    \mathcal{I'} = \{P'_i \in  \mathbb{R}^{h \times w \times \cpixel} \ | \ P'_i = f_\decoder(\mathcal{Z}'_i), 1 \leq i \leq M 
 \}.
\end{equation}
Herein, $P'_i$ represents the $i^{th}$ tile decoded from its corresponding denoised latent tile.

However, a seaming problem emerges when any two tiles in the set are disjoint, as depicted in Figure~\ref{tiled_diffusion}(a). To tackle this, we implement overlaps between neighboring tiles that share common pixels (Figure 3(b)). While increasing explicit overlap can effectively mitigate this issue, it substantially escalates the denoising time. As a consequence, the inference time quadratically increases with the growth in overlapping patches. Indeed, it's practically significant to strike a balance between inference time and the amount of overlap.

\noindent\textbf{Implicit overlap in tiled sampling.} 
To speed up the inference time while avoiding the seaming problem, we propose an implicit overlap in tiled sampling.
As depicted in Figure~\ref{tiled_diffusion}(c),
the magnified image is divided into $L$ non-overlapping tiles and we keep the quantity of disjoint noisy latent variables constant during the reverse sampling process.
Prior to each sampling step, we apply a random offset to each tile, effectively splitting $\mathcal{Z}$ into two components: $\mathcal{Z}^s$ (the shifted region with tiling) and $\mathcal{Z}^c$ (the constant region without tiling). This can be mathematically represented as $\mathcal{Z} = \mathcal{Z}^s \cup \mathcal{Z}^c$. 
Take note that at the initial time step, $\mathcal{Z}^c = \O$.
At each sampling, the shifted part, $\mathcal{Z}^s$, is a collection of $L$ disjoint tiles, denoted as $\mathcal{Z}^s = \{ \mathcal{Z}^s_i \ | \ 1 \leq i \leq L \}$. Here, each $\mathcal{Z}^s_i$ symbolizes a shifted tile.
The shifted portion, $\mathcal{Z}^s$, comprises $L$ disjoint tiles that change dynamically throughout the sampling process. Within this segment, each tile is expressed as $\mathcal{Z}^s_{i,x,y} = \mathcal{Z}_{y_i + \Delta y_i, x_i + \Delta x_i}$ for $1 \leq i \leq L$. Here, $\Delta x_i$ and $\Delta y_i$ denote the random offsets for tile $\mathcal{Z}^s_i$ implemented in the preceding step.
As for the constant section without tiling, denoted as $\mathcal{Z}^c$, the pixel value is sourced from the corresponding latent variable in the previous sampling step. It is worth noting that after each time step, $\mathcal{Z}^c$ is non-empty, symbolically represented as $\mathcal{Z}^c \neq \O$.
This approach ensures implicit overlap during tiled sampling, effectively solving the seaming issue.

\newcommand{\cmark}{\ding{51}}%
\newcommand{\xmark}{\ding{55}}%

\newcommand{\yuanfang}{\color{red}}%

\begin{figure*}
    \centering
    \includegraphics[width=1.0\linewidth]{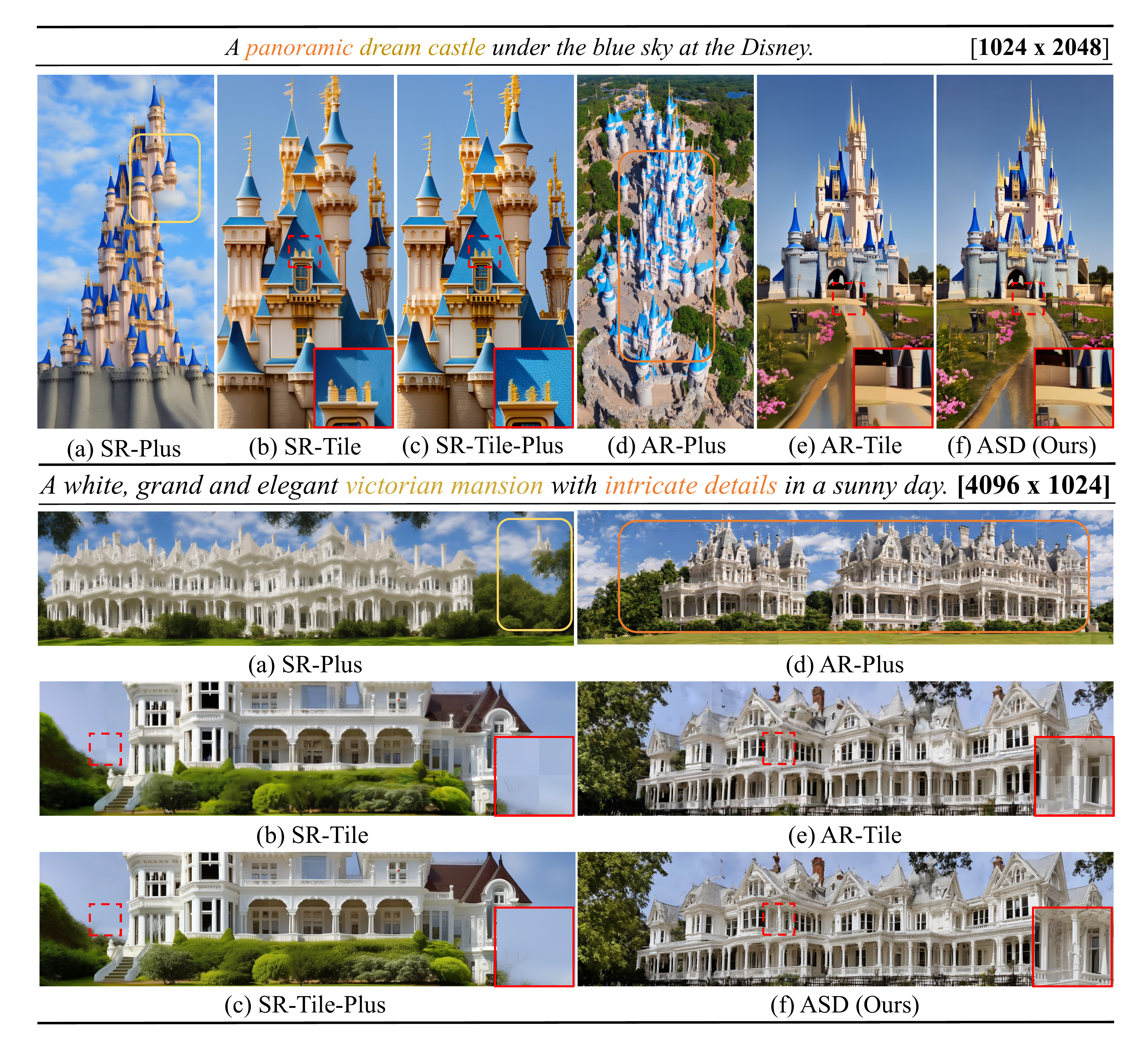}
    \vspace{-9.0mm}
    \caption{\textbf{Qualitative comparison of our proposed ASD method with other baselines}, including (a)~\textbf{SR-Plus}, (b)~\textbf{SR-Tile}, (c)~\textbf{SR-Tile-Plus}, (d)~\textbf{AR-Plus}, (e)~\textbf{AR-Tile} and (f)~our proposed~\textbf{\textit{ASD}}. The {\color{myyellow}{yellow}} box indicates the resolution-induced poor composition. The {\color{orange}{orange}} box indicates better composition. The {\color{red}{red}} solid line box is the zoom-in of the red dashed line box, aiming to inspect if there are any seaming issues. Our ASD outperforms others in both composition quality and inference time.}
    \label{fig:baseline_comparison}
\end{figure*}

\begin{table*}
\centering
\caption{\label{tab:baseline_comparison}\textbf{Quantitative evaluation against baselines}. (a)~\textbf{SR-Plus}, (b)~\textbf{SR-Tile}, (c)~\textbf{SR-Tile-Plus}, (d)~\textbf{AR-Plus}, (e)~\textbf{AR-Tile} and (f)~our ~\textbf{\textit{ASD}}. 
`S' and `A' denote single and arbitrary ratios, respectively. 
All tests run on a 32G GPU. Notably, under the same GPU memory, our ASD achieves at least 9$\times$ higher resolution than the original SD model.
\vspace{-1.0em}}
\begin{footnotesize}
  \begin{adjustbox}{max width=\linewidth}
  \begin{tabular}{c|cccccccccccc}
    \toprule
    \multirow{2}{*}{Exp.} &\multicolumn{1}{c}{\textbf{Stage-I}}&  &\multicolumn{2}{c}{\textbf{Stage-II}} & & 
    \multicolumn{3}{c}{\textbf{Capability}} & & \multicolumn{3}{c}{\textbf{MM-CelebA-HQ}} \\
    \cmidrule{2-2} \cmidrule{4-5} \cmidrule{7-9} \cmidrule{11-13} 
    &Ratio & &Tile &Overlap & &Composition &Max Resolution &Seam & &FID $\downarrow$ & IS $\uparrow$ & CLIP $\uparrow$ \\
    \cmidrule{1-2} \cmidrule{4-5} \cmidrule{7-9} \cmidrule{11-13}
     (a)&$\mathbf{S}$ & &\XSolidBrush &\XSolidBrush & 
        &Poor &$2048^2$  &N & & 118.83 & 2.11 & 27.22 \\
     (b)&$\mathbf{S}$ & &\Checkmark &\XSolidBrush & 
        &Poor &$18432^2$ &Y & & 111.96  {\footnotesize (\textcolor{color3}{- 6.87})} & 2.46 {\footnotesize (\textcolor{color2}{+ 0.35})} & 27.46 {\footnotesize (\textcolor{color2}{+ 0.24})}\\
     (c)&$\mathbf{S}$ & &\Checkmark &\Checkmark & 
        &Poor &$18432^2$  &N & & 111.06 {\footnotesize (\textcolor{color3}{- 7.77})}& 2.53 {\footnotesize (\textcolor{color2}{+ 0.42})}& 27.55 {\footnotesize (\textcolor{color2}{+ 0.33})}\\
     (d)&$\mathbf{A}$ & &\XSolidBrush &\XSolidBrush & 
        &Excellent &$2048^2$ &N & & 92.80 {\footnotesize (\textcolor{color3}{- 26.03})} & 3.97 {\footnotesize (\textcolor{color2}{+ 1.86})}& 29.15 {\footnotesize (\textcolor{color2}{+ 1.93})}\\
     (e)&$\mathbf{A}$ & &\Checkmark &\XSolidBrush & 
        &Excellent &$18432^2$ &Y & &85.66 {\footnotesize (\textcolor{color3}{- 33.17})} & 3.98 {\footnotesize (\textcolor{color2}{+ 1.87})} &29.17 {\footnotesize (\textcolor{color2}{+ 1.95})}\\
     (f)& \cellcolor{mygray}$\mathbf{A}$ &\cellcolor{mygray} &\cellcolor{mygray}\Checkmark &\cellcolor{mygray}\Checkmark &\cellcolor{mygray} 
     &\cellcolor{mygray}Excellent &\cellcolor{mygray}$18432^2$ 
     &\cellcolor{mygray}N 
     &\cellcolor{mygray} 
     &\cellcolor{mygray}\textbf{85.34} {\footnotesize (\textcolor{color3}{- 33.49})}
     &\cellcolor{mygray}\textbf{4.04} {\footnotesize (\textcolor{color2}{+ 1.93})}
     &\cellcolor{mygray}\textbf{29.23} {\footnotesize (\textcolor{color2}{+ 2.01})} \\
\bottomrule
    \end{tabular}
    \end{adjustbox}
\end{footnotesize}
\end{table*}

\section{Experiments}

\subsection{Experimental Settings}

\noindent\textbf{Datasets.} 
%
The ARAD of our ASD is trained on a subset of LAION-Aesthetic~\cite{laionaesthetics} with 90k text-image pairs in different aspect ratios. It is evaluated on MA-LAION-COCO with 21,000 images across 21 ratios (selecting from LAION-COCO~\cite{laioncoco}), and MA-COCO built from MS-COCO~\cite{Lin2014MicrosoftCC} containing 2,100 images for those ratios. 
A test split of MM-CelebA-HQ~\cite{xia2021tedigan}, consisting of 2,824 face image pairs in both low and high resolutions, is employed to evaluate our FSTD
and whole pipeline.

\noindent\textbf{Implementation Details.}
Our proposed method is implemented in PyTorch \cite{paszke2019pytorch}. A multi-aspect ratio training method is leveraged to finetune ARAD (using LoRA \cite{lora}) for 10,000 steps with a batch size of 8. We use Adam~\cite{kingma2014adam} as an optimizer and the learning rate is set to 1.0e-4. Our FSTD (the second stage model) is training-free and is built upon StableSR~\cite{wang2023exploiting}. During inference, DDIM sampler \cite{song2020denoising} of 50 steps is adopted in ARAD to generate the image according to the user-defined aspect ratio. In the second stage, we follow StableSR to use 200 steps DDPM sampler~\cite{NIPS2020_diffusion} for FSTD.


\noindent \textbf{Evaluation metrics.} 
For benchmarks, we employ common perceptual metrics to assess the generative text-to-image models, including FID \cite{NIPS2017_8a1d6947}, IS \cite{NIPS2016_8a3363ab} and CLIP \cite{NIPS2021_8a3363ab}.
IS correlates with human judgment, important to evaluate the metric on a large enough number of samples. 
FID captures the disturbance level very well and is more consistent with the noise level than the IS.
CLIP score is used to measure the cosine similarity between the text prompt and the image embeddings.
Besides, the extra metrics (\textit{e.g.,} PSNR, SSIM \cite{wang2004image} and LPIPS \cite{Zhang_2018_CVPR}) are employed to assess the super-resolution ability of the second stage of our ASD.
PSNR and SSIM scores are evaluated on the luminance channel in the YCbCr color space.
LPIPS quantifies the perceptual differences between images.

\subsection{Baseline Comparisons}
Based on state-of-the-art diffusion models, we build the following six baselines for comparison. 
\begin{itemize}
\item \textbf{SR-Plus:} employs SD 2.1 \cite{Rombach_2022_CVPR} for the direct synthesis of text-guided images with varying sizes.
\item \textbf{SR-Tile:} utilizes SD 2.1 for initial image generation, magnified using StableSR \cite{wang2023exploiting} with a non-overlap in tiled sampling\cite{jiménez2023mixture}.
\item \textbf{SR-Tile-Plus:} A two-stage method that initiates with SD 2.1 \cite{Rombach_2022_CVPR} and refines the output using our proposed FSTD, facilitating the synthesis of images of arbitrary dimensions.
\item \textbf{AR-Plus:} deploys our proposed ARAD model for direct, text-driven image synthesis across a spectrum of sizes.
\item \textbf{AR-Tile:} commences with our ARAD model for initial image generation, followed by magnification via StableSR employing a non-overlap in tiled sampling.
\item \textbf{ASD:} is our proposed novel framework, integrating ARAD in Stage I and FTSD in Stage II, designed to synthesize images with customizable dimensions.
\end{itemize}

\noindent \textbf{Quantitative evaluation.}
As reported in Table~\ref{tab:baseline_comparison}, our proposed ASD method consistently outperforms the baseline methods. Specifically, our ASD model shows a 33.49 reduction in FID score compared to (a) SR-Plus, and an increase of 1.92 and 2.01 in IS and CLIP scores, respectively.
On a 32GB GPU, SR-Plus fails to synthesize images exceeding 2048$^2$ resolution. In contrast, our ASD effectively mitigates this constraint, achieving at least 9$\times$ higher resolution than SR-Plus under identical hardware conditions.
Additionally, we also have the following observations:
\textbf{(i)} Utilizing multi-aspect ratio training results in notable improvements across various comparisons, specifically reducing FID  scores from 118.83 to 92.80 in (a)-(d), 111.96 to 85.66 in (b)-(e), and 111.06 to 85.34 in (c)-(f).
\textbf{(ii)} Introducing a tiled algorithm at the second stage enables the generation of images with unlimited resolution, while simultaneously enhancing performance, \textit{e.g.,} FID scores improve from 92.80 to 85.66 when comparing (a)-(b) and (d)-(c).
\textbf{(iii)} Implementing overlap in tiled sampling effectively addresses the seaming issue, as evidenced by the comparisons 
between (b)-(c) and (e)-(f).

\noindent \textbf{Qualitative comparison.} 
As depicted in Fig.~\ref{fig:baseline_comparison}, the images synthesized by ASD exhibit superior composition quality (\textit{e.g.} proper layout) when compared to other baseline methods.
Additionally, ASD can generate 4K HD images that are not only well-composed but also free from seaming artifacts. 
Specifically, when guided by a text description, the AR-Plus method is observed to generate a more complete castle than SR-Plus, as demonstrated in Fig.\ref{fig:baseline_comparison}(a) \textit{vs.} Fig.\ref{fig:baseline_comparison}(d).
Compared with SR-Plus, AR-Tile can produce realistic images but is hindered by the presence of seaming issues (see Fig.~\ref{fig:baseline_comparison}(e)). 
In contrast, Fig.~\ref{fig:baseline_comparison}(f) shows that our ASD successfully eliminates seaming artifacts and ensures the production of well-composed images, while minimizing GPU memory usage.

\begin{table}
\centering
\caption{\label{tab:stage-1} \textbf{Comparison of our ARAD and other diffusion-based approaches.} We compare their compositional ability to handle the synthesis of images across 21 different sizes. 
\vspace{-1.0em}}
  \begin{adjustbox}{max width=\linewidth}
  \begin{tabular}{cccccccccc}
    \toprule
    \multicolumn{2}{c}{\multirow{2}{*}{\textbf{Method}}} & &\multicolumn{3}{c}{\textbf{MA-LAION-COCO}} & & \multicolumn{3}{c}{\textbf{MA-COCO}}\\
     \cmidrule{4-6} \cmidrule{8-10} 
      &  & & FID $\downarrow$ & IS $\uparrow$ & CLIP $\uparrow$ & & FID $\downarrow$ & IS $\uparrow$ & CLIP $\uparrow$  \\
    \cmidrule{1-2} \cmidrule{4-6} \cmidrule{8-10}
    \multicolumn{2}{l}{SD$_{2.1}$}   &  & 14.32 & 31.25 & 31.92 & & 42.50 & 30.20 & 31.63 \\
    \multicolumn{2}{l}{MD$_{2.1}$}   &  & 14.57 & 28.95 & 32.11 & & 43.25 &  28.92 & 30.92 \\
   \rowcolor{mygray} \multicolumn{2}{l}{ARAD}  &  & \textbf{13.98}  & \textbf{34.03} & \textbf{32.60} & & \textbf{40.28} &  29.77 & \textbf{31.87} \\
\bottomrule
    \end{tabular}
    \end{adjustbox}
\vspace{-5.0mm}
\end{table}

\begin{figure}
    \centering
    \includegraphics[width=1.0\linewidth]{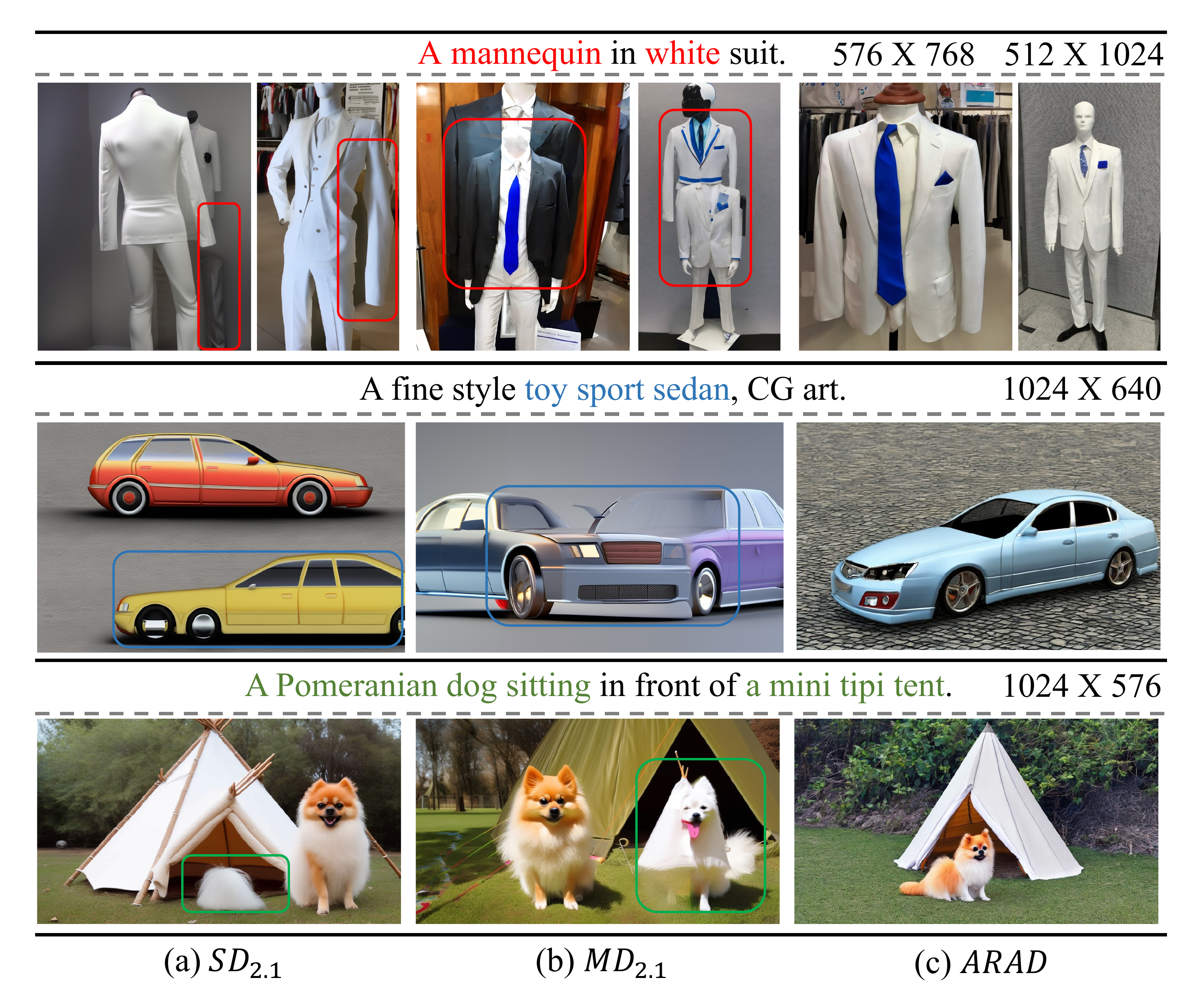}
    \vspace{-4.0mm}
    \caption{\textbf{Comparison of visual results.} Composition quality of the text-to-image synthesis using (a) SD$_{2.1}$, a stable diffusion 2.1, (b) MD$_{2.1}$, a multi-diffusion based on SD 2.1, and (c) our ARAD. Color boxes indicate poor composition.}
    \label{fig:stage-1}
    \vspace{-6.0mm}
\end{figure}

\begin{table}
\centering
\caption{\label{tab:stage-1-ratio}\textbf{Performance on ARAD trained on the various types of aspect ratios.} ``All'' denotes the 9 aspect ratios.\vspace{-1.0em}}
  \begin{adjustbox}{max width=\linewidth}
  \begin{tabular}{ccccccccc}
    \toprule
    \multicolumn{1}{c}{\multirow{2}{*}{\textbf{Types}}} & &\multicolumn{3}{c}{\textbf{MA-LAION-COCO}} & & \multicolumn{3}{c}{\textbf{MA-COCO}}\\
     \cmidrule{3-5} \cmidrule{7-9} 
      &  & FID $\downarrow$ & IS $\uparrow$ & CLIP $\uparrow$ & & FID $\downarrow$ & IS $\uparrow$ & CLIP $\uparrow$  \\
    \cmidrule{1-1} \cmidrule{3-5} \cmidrule{7-9}
    \multicolumn{1}{c}{3}   &  & 14.36 & 32.53 & 32.38 & &41.28 &29.58 &31.71 \\
    \multicolumn{1}{c}{5}   &  & 14.10 & 33.61 & 32.58 & &40.25 &29.63 &31.80 \\
    \rowcolor{mygray} \multicolumn{1}{c}{All} &  & \textbf{13.98} & \textbf{34.03} & \textbf{32.60} & &\textbf{40.28} &\textbf{29.77} &\textbf{31.87} \\
\bottomrule
    \end{tabular}
    \end{adjustbox}
\vspace{-1.0em}
\end{table}

\subsection{ARAD Analysis}
To verify the superiority of our proposed ARAD in addressing resolution-induced poor composition issues, we conduct the ablation study, specifically at the initial stage.

\noindent\textbf{Impact of ARAD.}
Table~\ref{tab:stage-1} highlights the performance of ARAD, showing improvements of 13.98, 34.03, and 32.60 in FID, IS, and CLIP, respectively, on MA-LAION-COCO over original SD 2.1 and MultiDiffusion \cite{bar2023multidiffusion} (MD$_{2.1}$). This superiority is further illustrated in Fig.~\ref{fig:stage-1}. While SD$_{2.1}$ and MD$_{2.1}$ exhibit composition problems, our ASD produces images that are consistent with user-defined textual descriptions. For example, MD$_{2.1}$ incorrectly generates two overlapped blue suits from a prompt for a white suit, a mistake not present in our ASD's results.

\noindent\textbf{Influence on the number of aspect ratios.}
Table~\ref{tab:stage-1-ratio} reveals the model's performance across various aspect ratios. The data shows that increasing the number of aspect ratios in the training dataset improves performance, with FID scores falling from 14.36 to 13.98.
A comparison between 3 and 5 aspect ratios highlights a significant improvement, as the FID score drops from 14.36 to 14.10. Further increasing the aspect ratios continues this trend, reducing the FID score to 13.98. This pattern emphasizes the importance of aspect ratios in enhancing model performance.

\begin{figure}
    \centering
    \includegraphics[width=1.0\linewidth]{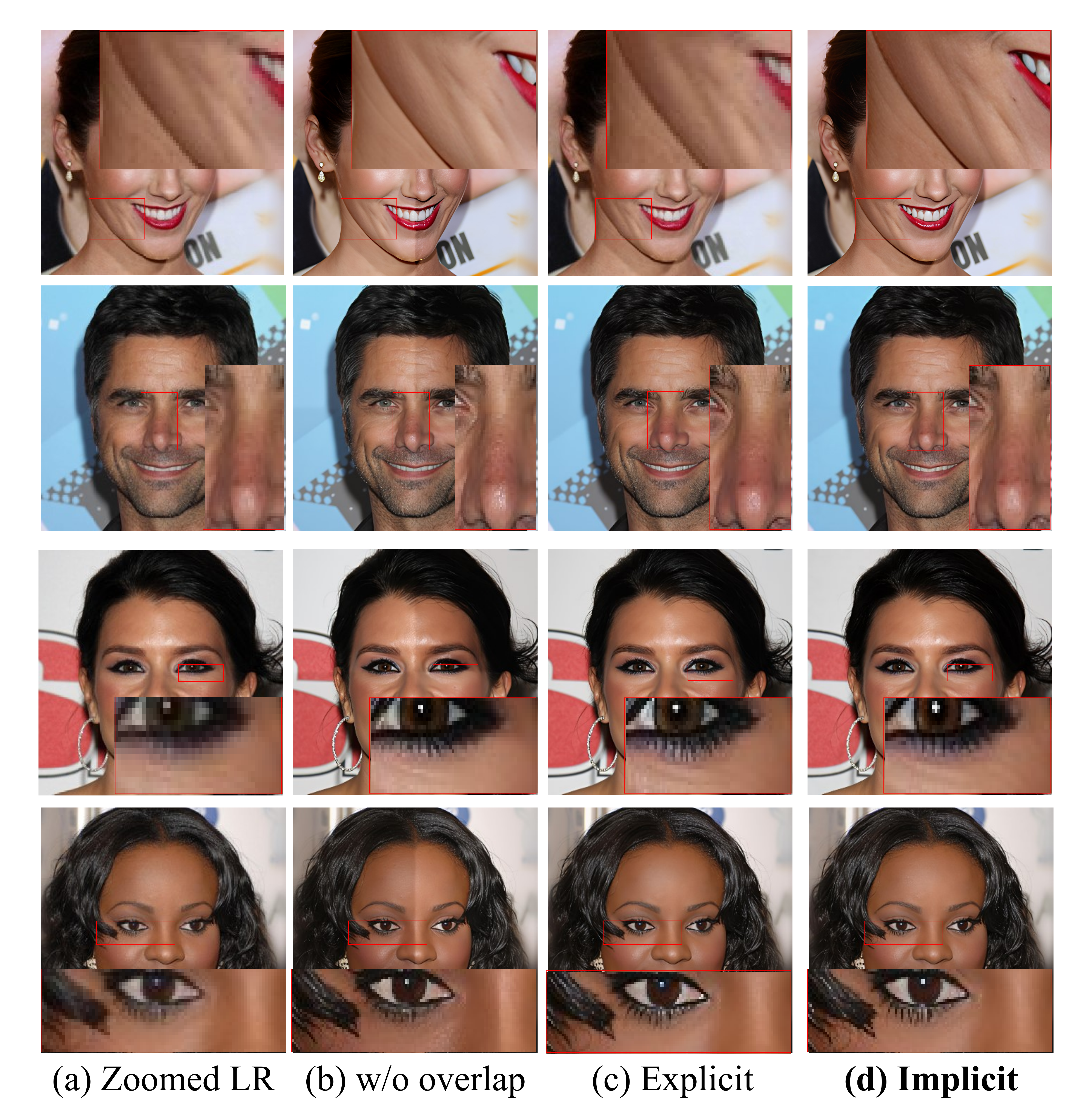}
    \vspace{-8.0mm}
    \caption{\textbf{The super-resolution results of $\times$4 for different methods.} We visually compare the (a) Zoomed LR (bicubic method), tiled diffusion with (b) non-overlap and (c) explicit overlap tiles; and (d) our FSTD which uses implicit overlap in tiled sampling. \textbf{Notably, (d) is \textit{$2\times$ faster} than (c)}.}
    \label{fig:stage-2}
    \vspace{-8.0mm}
\end{figure}
\begin{table}
\centering
\caption{\label{tab:stage-2} \textbf{The versatility of tiled sampling in FSTD.} We conduct ablation on the MM-CelebA-HD benchmark. ``w/o'', ``explicit'', and ``implicit'' describe non-overlapping, explicit, and implicit overlap in tile sampling respectively. ``fixed'', and ``random'' refer to different tile offset strategies. Here, the overlap of two adjacent tiles is 32$\times$32. \vspace{-1.0em}}
  \begin{adjustbox}{max width=\linewidth}
  \begin{tabular}{lcccccc}
    \toprule
    \textbf{Method} && \multicolumn{4}{c}{\textbf{MM-CelebA-HQ}}  & \textbf{Time} \\
     \cmidrule{3-6}
    Overlap \& Offset && PSNR$\uparrow$ & SSIM$\uparrow$ & LPIPS$\downarrow$ & FID$\downarrow$  & per frame \\
    \midrule
    w/o overlap && 26.89 & 0.76 & 0.09 & 22.80 & 75.08s \\
    explicit    && 27.49 & 0.76 & 0.09 & 24.15 & 166.8s \\
    \cmidrule(lr){1-1} \cmidrule(lr){3-3} \cmidrule(lr){4-4} \cmidrule(lr){5-5} \cmidrule(lr){6-6} \cmidrule(lr){7-7} 
    implicit \& fixed && 26.83 & 0.75 & 0.08 & 21.37 & \textbf{75.01s} \\
    \rowcolor{mygray} implicit \& random && \textbf{27.53} & 0.76 & 0.08 & 22.25 & 75.19s \\
\bottomrule
    \end{tabular}
    \end{adjustbox}
    \vspace{-1.0em}
\end{table}

\subsection{FSTD Analysis}
Although we have proved the effectiveness of the proposed FSTD in Fig.~\ref{fig:baseline_comparison} and Table~\ref{tab:baseline_comparison}, we now explore its design in the image super-resolution performance on MM-CelebA-HD. Table~\ref{tab:stage-2} report the ablation study on the versatility of tiled sampling; see more details in the supplementary material.

\noindent\textbf{Importance of tiles with overlap.} 
The first two lines from Table~\ref{tab:stage-2} reveal a comparison between the perceptual performance of explicit overlap and non-overlap in tiled sampling. Specifically, the explicit overlap exhibits superior performance (\textit{e.g.,} 27.49 \textit{vs.} 26.89 on PSNR). However, non-overlap tiled sampling offers an approximately 2$\times$ faster inference time compared to the explicit overlap. Despite this advantage in speed, Fig.~\ref{fig:stage-2}(b) clearly exposes the seaming problem associated with non-overlap tiled sampling, highlighting the trade-off between performance and efficiency.

\noindent\textbf{Implicit vs. explicit overlap.}
An analysis of the results presented in Table~\ref{tab:stage-2} and Fig.\ref{fig:stage-2}(c)-(d) confirms that the use of implicit overlap in tiled sampling yields the best performance across both perceptual metrics and visual representation. Further examination of the last column in Table\ref{tab:stage-2} demonstrates that the inference time for implicit overlap in tiled sampling is nearly equivalent to that of tiling without overlap. Moreover, the implementation of implicit overlap successfully reduces the inference time from 166.8s to approximately 75.0s. This ablation study validates the superiority of our proposed FSTD method, accentuating its capacity to achieve an optimal balance between performance quality and inference time.

\noindent\textbf{Effect of various offset strategies.}
The last two lines of Table~\ref{tab:stage-2} demonstrate the advantage of using a random offset in implicit overlap tiled sampling.
Specifically, when comparing the fixed and random offset methods in implicit overlap, the random offset yields a PSNR value of 27.53, outperforming the fixed offset, which registered at 26.83. The results for other perceptual metrics and visual performance indicators are found to be nearly identical, further emphasizing the preference for a random offset in this context.






\section{Conclusion} 

In this study, we address the challenge of resolution-induced poor composition in creating high-fidelity images from any text prompt. We propose \emph{Any Size Diffusion} (ASD), a method consisting of ARAD and FSTD. Trained with multi-aspect ratio images, ARAD generates well-composed images within specific sizes. FSTD, utilizing implicit overlap in tiled sampling, enlarges previous-stage output to any size, reducing GPU memory consumption. Our ASD is validated both quantitatively and qualitatively on real-world scenes, offering valuable insights for future work.



\twocolumn[ 
\begin{@twocolumnfalse} 
  \vspace{15mm}
  \vspace*{\fill}
  \begin{center}
    \Huge\bfseries
    Appendix
  \end{center}
  \vspace*{\fill}
  \vspace{15mm}
\end{@twocolumnfalse}
]

\section{Details of Our ASD Methodology}
In this section, we present a comprehensive analysis of our proposed Any Size Diffusion (ASD) pipeline. We first delineate the multi-aspect ratio training strategy implemented in Stage-I. Subsequently, we provide a thorough examination of the implicit overlap characteristics inherent to the tiled sampling approach adopted in Stage II.

\paragraph{Stage-I: Any Ratio Adaptability Diffusion (ARAD)}
Algorithm~\ref{alg:asad} presents a detailed description of our proposed multi-aspect ratio training strategy. 
We establish nine pre-defined key-value pairs, where each key represents a distinct aspect ratio and each value corresponds to a specific size. The training images, which vary in size, are processed according to this multi-aspect ratio strategy, as specified in lines 4-12 of Algorithm~\ref{alg:asad}.

\vspace{-1.0em}
\begin{algorithm}
\caption{Multi-Aspect Ratio Training Strategy}
\label{alg:asad}
\LinesNumbered
\textbf{Input: }Image $\mathcal{I} \in \mathbb{R}^{H \times W\times 3}$, 
          Ratio-Size Dictionary $D_{r\rightarrow s} = \{r_1:s_1, r_2:s_2, \cdots, r_9: s_9\}$ \;
\textbf{Output: } Resized Image for Model Training \;

Compute image ratio $r = \frac{H}{W}$ \;
Initialize minValue = $\infty$ \; 
Initialize minIndex = 0 \;

\For{\rm{each ratio} $r_i$ \rm{in} $\{r_1, r_2, \cdots, r_9\}$} 
{
     Compute distance = $|r - r_i|$ \;
     \If{distance $\leq$ minValue}
     {
        minValue = distance \;
        minIndex = i
     }
     $i$ = $i + 1$ \;
}
\textbf{end} \;
Retrieve $r_m$, $s_m$ from $D_{r\rightarrow s}$ using minIndex \;
Resize image $\mathcal{I}$ to $s_m$ for model training
\end{algorithm}

\vspace{-1.0em}
\begin{algorithm}
\caption{Procedures for our proposed FSTD.}
\label{alg:fstd}
\LinesNumbered
\KwIn{an image $\mathcal{I} \in \mathbb{R}^{H \times W\times 3}$}
\KwOut{an image $\mathcal{I'} \in \mathbb{R}^{H \times W\times 3}$}
\texttt{\\}
$\blacktriangleright$ \textbf{Step 1: Tiled Sampling Preparation} \\
Divided the input image $\mathcal{I}$ into a set of $M$ disjoint tiles: $\{P_i^{h \times w \times 3} \ | \ 1 \leq i \leq M \} $ \;
\For{ \rm {each tile} $P_i$ \rm{in} \{$P_1, P_2,\cdots, P_M$\}} 
{
    \textcolor{gray}{/* \textit{add a random noise to latent} */} \\
    $L_i$ = $\encoder(P_i) + \epsilon_i$ \;
}
\textbf{end} \;
Hence, we have $M$ \textbf{disjoint} latents \{$L_1, L_2,\cdots, L_M$\} \;
\texttt{\\}
$\blacktriangleright$ \textbf{Step 2: Implicit Overlap Tiled Sampling} \\
Suppose that $\mathcal{Z}^s = \{L_1, L_2,\cdots, L_M\}$ and $\mathcal{Z}^c = \O$, and $\mathcal{Z} = \mathcal{Z}^s \cup \mathcal{Z}^c$, and the total step is $T$ \;
\texttt{\\}
Initialize $\mathcal{Z}_T^s = \{L_1, L_2,\cdots, L_M\}$ \; 
Initialize $\mathcal{Z}_T^c = \O$ \;
Initialize $\mathcal{Z}_T = \mathcal{Z}_T^s \cup \mathcal{Z}_T^c$\;
\texttt{\\}
\textcolor{red}{$\bigstar$ \textbf{Implicit Overlaps for Tiles}} \\
\For{ \rm {each time step} $t$ \rm{in} $\{T-1,\cdots, 0\}$} 
{
    Set a random offset $(\Delta x, \Delta y)$\;
    \For{ \rm {each tile} $L_i$ \rm{in} $\{L_1, L_2,\cdots, L_M\}$}
    {
        $L_{i,x_i,y_i}$ = $L_{i, x_i + \Delta x_i, y_i + \Delta y_i}$ \;
    }
    \textbf{end} \;
    \textcolor{gray}{/*  Update $\mathcal{Z}^s$ and $\mathcal{Z}^c$ */} \\
    $\mathcal{Z}_t^s = \{L_{1,x_1,y_1}, L_{2,x_2,y_2},\cdots, L_{M,x_M,y_M}\}$ and the number of the tiles keeps constant \; 
    $\mathcal{Z}_t^c$ = $\mathcal{Z}_{t+1} \setminus \mathcal{Z}_t^s$ \;
    \textcolor{gray}{/*  Denoise updated latents */} \\
    \For{ \rm {each tile} $L_{i,x_i,y_i}$ \rm{in} $\mathcal{Z}_t^s$}
    {
        Apply UNetModel to denoise ($L_{i,x_i,y_i}$ ) \;
    }
    \textbf{end} \;
    $\mathcal{Z}_t$ = $\mathcal{Z}_t^s \cup \mathcal{Z}_t^c$  and $\mathcal{Z}_t^c \neq \O$\;
}
\textbf{end} \;
\texttt{\\}
After denoising for $T$ timesteps, $\mathcal{Z}' = \mathcal{Z}_0$ \;
Consequently, we have $\mathcal{Z}'$ and $\mathcal{I'} = \decoder(\mathcal{Z}')$.
\end{algorithm}

\paragraph{Stage-II: Fast Seamless Tiled Diffusion (FSTD).}
Algorithm \ref{alg:fstd} outlines the step-by-step procedure of our proposed Fast Seamless Tiled Diffusion (FSTD) technique, designed to efficiently upscale a low-resolution image to a target resolution of 
$H \times W$.
Initially, as delineated from line 1 to line 7 in Algorithm \ref{alg:fstd}, we partition the input low-resolution image into $M$ non-overlapping tiles, with each being represented as a separate latent variable \{$L_1, L_2,\cdots, L_M$\}.
These latent variables collectively form a set $\mathcal{Z}$, as described in line 8.
In particular, $\mathcal{Z}$ is composed of two distinct components, as assumed in line 11: the shifted region $\mathcal{Z}^s$ and the constant region $\mathcal{Z}^c$. The former is designated for processing through tiled denoising, while the latter remains excluded from the tiling process at the current time step.
At the initial $T^{th}$ time step, as illustrated in lines 13-15, $\mathcal{Z}^s$ is initialized as \{$L_1, L_2,\cdots, L_M$\}, and $\mathcal{Z}^c$ is initialized as an empty set. This is due to the initial offsets of each tile being set to zero.
At the heart of the algorithm, as exhibited from lines 17 to 31, is a novel mechanism termed `Implicit Overlap in Tiled Sampling'. This mechanism is conceived to significantly reduce both inference time and GPU memory consumption compared to the non-tiled sampling method used in stable diffusion processes. For each time step, the algorithm randomly offsets the positions of these latent variables while maintaining their quantities invariant. This results in an updated shifted region, denoted as $\mathcal{Z}_t^s \in  \{L_{1,x_1,y_1}, L_{2,x_2,y_2},\cdots, L_{M,x_M,y_M}\}$, and a novel constant region $\mathcal{Z}_t^c \in \mathcal{Z}_{t+1} \setminus \mathcal{Z}_t^s$. 
Notably, from the second denoising step onwards, $\mathcal{Z}_t^c$ becomes a non-empty set, with each pixel within this constant region retaining the same value as the corresponding pixel in the preceding denoised latent variables.
After $T$ time steps, the iterative procedure obtains a new denoised latent set, denoted as $\mathcal{Z}'$. In the final stage, the algorithm decodes this latent set $\mathcal{Z}'$ back into an image of the user-defined size $H \times W \times 3$, thereby producing the final super-resolved output image.

\section{Implementation Details}
This section elaborates on the implementation details associated with the multi-aspect ratio training approach, and delineate  configurations integral to the tiled sampling strategy.

\paragraph{Multi-aspect ratio training.} We establish a set of nine distinct aspect ratios, each associated with a corresponding size specification, as enumerated in Table~\ref{tab:ratio-size}. Prior to the commencement of training, images are systematically resized to a predetermined dimensionality, as prescribed by Algorithm~\ref{alg:asad}. Subsequently, images exhibiting identical aspect ratios are agglomerated into a stochastic batch, which is then utilized as the input for model training.

\paragraph{Explicit and implicit overlaps in tiled sampling.} 
In the context of tiled sampling, we explore two distinct strategies: explicit and implicit overlap configurations. Notably, in both strategies, the dimensions of the input image and tiles can be parameterized by the user.
For the explicit overlap configuration, we mathematically formulate the relationship between the number of tiles and their overlaps as follows:
\begin{equation}
\label{eq:tile-overlap-relation}
N_{tiles} = \lfloor \frac{W_{image}}{(W_{tile} - overlap)} \times \frac{H_{image}}{(H_{tile} - overlap)} \rfloor,
\end{equation}
where $W_{image}$ and $H_{image}$ represent the width and height of the input image, respectively, while $W_{tile}$ and $H_{tile}$ denote the corresponding dimensions of each tile.
In contrast, the implicit overlap strategy conventionally configures the size of each tile to $512 \times 512$ with zero overlaps between adjacent tiles. Moreover, the spatial dimensions of latent variables are $64 \times 64$. To introduce a form of overlap, we employ a random offset strategy, designed to control the translational shifts of these tiled latent representations, thereby achieving implicit overlap of tiles.
Specifically on the MM-CelebA-HQ~\cite{xia2021tedigan} benchmark, we utilize high-resolution images with dimensions of $1024 \times 1024$ alongside corresponding low-resolution images sized $256 \times 256$.


\begin{table}
\centering
\caption{\label{tab:ratio-size} \textbf{Dictionary of nine pre-defined ratio-size pairs.} Each ratio corresponds to a pre-defined image size.\vspace{-1.0em}}
  \begin{adjustbox}{max width=\linewidth}
  \begin{tabular}{lcccc}
    \toprule
    \textbf{\#} & & \textbf{Ratio}  & & \textbf{Size} \\
    \midrule
    1 & & 1.0000 && $512 \times 512$  \\
    2 & & 0.7500 && $576 \times 768$  \\
    3 & & 1.3330 && $768 \times 576$  \\
    4 & & 0.5625 && $576 \times 1024$  \\
    5 & & 1.7778 && $1024 \times 576$  \\
    6 & & 0.6250 && $640 \times 1024$  \\
    7 & & 1.6000 && $1024 \times 640$  \\
    8 & & 0.5000 && $512 \times 1024$  \\
    9 & & 2.0000 && $1024 \times 512$  \\
\bottomrule
    \end{tabular}
    \end{adjustbox}
\end{table}

\section{More Ablation Study}
In this section, we present extensive ablation studies to conduct an in-depth analysis of our proposed model. Our analysis is divided into two primary components: the first component focuses specifically on our Fast Seamless Tiled Diffusion (FSTD) technique, while the second component encompasses a comprehensive evaluation of the entire pipeline, which we denote as Any Size Diffusion (ASD).

\paragraph{Impact of the random offset range in FSTD.}
In Table~\ref{tab:stage-2-offset}, we present a detailed analysis, revealing that the range of the random offset has minimal influence on the super-resolution performance of images within the MM-CelebA-HQ dataset. Specifically, with offset ranges of 16, 32, and 48, the PSNR scores exhibit remarkable consistency, recording values of 27.51, 27.53, and 27.52, respectively. Furthermore, the inference times across these distinct offset ranges remain similarly uniform. This observation underscores the robustness of our approach, as it performs consistently well under varying offset parameters, thereby demonstrating its resilience to changes in this aspect of the configuration.

\begin{table}
\centering
\caption{\label{tab:stage-2-offset} \textbf{The Effect of different random offset ranges in FSTD.} We conduct ablation on the MM-CelebA-HD benchmark by upscaling the low-resolution images of sizes 256 x 256 to 1024x1024. The range of random offset is 16, 32 and 48. The row in {\color{gray}gray} is the default setting of our method.\vspace{-1.0em}}
  \begin{adjustbox}{max width=\linewidth}
  \begin{tabular}{lcccccc}
    \toprule
    \textbf{Offset} && \multicolumn{4}{c}{\textbf{MM-CelebA-HQ}}  & \textbf{Time} \\
     \cmidrule{3-6}
    range && PSNR$\uparrow$ & SSIM$\uparrow$ & LPIPS$\downarrow$ & FID$\downarrow$  & per image \\
    \midrule
    16 && 27.51 & 0.76 & 0.09 & 22.58 & 75.39s \\
    \rowcolor{mygray} 32    && 27.53 & 0.76 & 0.08 & 22.25 & 75.19s \\ 
    48 && 27.52  & 0.76 & 0.08 & 22.06 & 75.21s \\
\bottomrule
    \end{tabular}
    \end{adjustbox}
    \vspace{-1.0em}
\end{table}

\begin{table}
\centering
\caption{\label{tab:stage-2-overlap} \textbf{Performance comparison across various numbers of explicit tiles.}
We conduct ablation on the MM-CelebA-HD benchmark by upscaling the low-resolution images in 256 x 256 to 1024x1024. The number of tiles in total with respect to overlap could be calculated by Eq~\ref{eq:tile-overlap-relation}.\vspace{-1.0em}}
  \begin{adjustbox}{max width=\linewidth}
  \begin{tabular}{lccccccc}
    \toprule
    \multicolumn{2}{c}{\textbf{Method}} && \multicolumn{4}{c}{\textbf{MM-CelebA-HQ}}  & \textbf{Time} \\
    \cmidrule{1-2} \cmidrule{4-7}
    Overlap & $N_{tiles}$ && PSNR$\uparrow$ & SSIM$\uparrow$ & LPIPS$\downarrow$ & FID$\downarrow$  & per image \\
    \midrule
    w/o  & $16^2$ && 26.89 & 0.76 & 0.09 & 22.80 & \textbf{75.1s} \\
    16   & $21^2$ && 27.50 {\footnotesize (\textcolor{color2}{+ 0.61})} & 0.76 & 0.09 & 23.21 & 148.7s \\
    32   & $32^2$ && 27.49 & 0.76 & 0.09 & 24.15 & 166.8s \\
    \rowcolor{mygray} 48   & $64^2$ && \textbf{27.64} & 0.76 & 0.09 & 24.25 & 182.6s {\footnotesize (\textcolor{red}{+ 33.9})} \\
\bottomrule
    \end{tabular}
    \end{adjustbox}
    \vspace{-1.0em}
\end{table}

\paragraph{Influence on the number of tiles or overlap region in explicit overlap tiled sampling.}
Table~\ref{tab:stage-2-overlap} illustrates the trade-off between perceptual quality and computational efficiency in relation to tile overlap in image super-resolution. Notably, as the number of tiles increases, there is a corresponding improvement in perceptual metrics, although this comes at the cost of increased computational time. For instance, tiles with a 16-pixel overlap exhibit superior perceptual metrics compared to non-overlap tiles, yielding a notable improvement of 0.63 in PSNR score. However, this enhancement comes with a substantial increase in inference time, which increases from 75.1s to 148.7s. Further, compared to the results presented in the second row of the table, a tile overlap of 48 pixels yields a PSNR score improvement from 27.50 to 27.64, while incurring an additional inference time of 33.9s.

\begin{figure}
    \centering
    \includegraphics[width=1.0\linewidth]{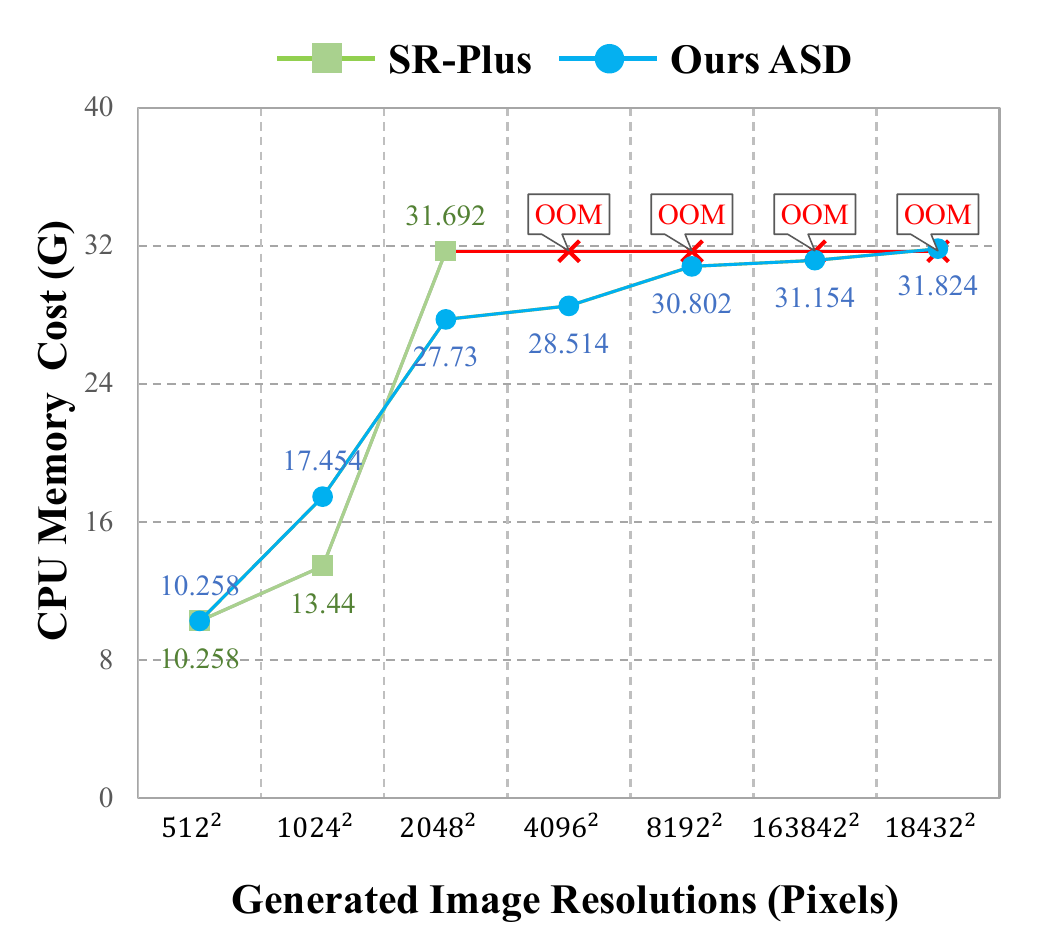}
    \vspace{-8.0mm}
    \caption{\textbf{Comparison of SR-Plus and ours ASD in terms of GPU memory cost (G) vs. image resolution (pixels).} SR-Plus is the original SD 2.1 model used to generate images of varying sizes. OOM is the abbreviation of an out-of-memory error. Experiments are conducted on a 32G GPU.\vspace{-1.0em}}
    \label{fig:resolutions}
\end{figure}

\begin{figure}
    \centering
    \includegraphics[width=1.0\linewidth]{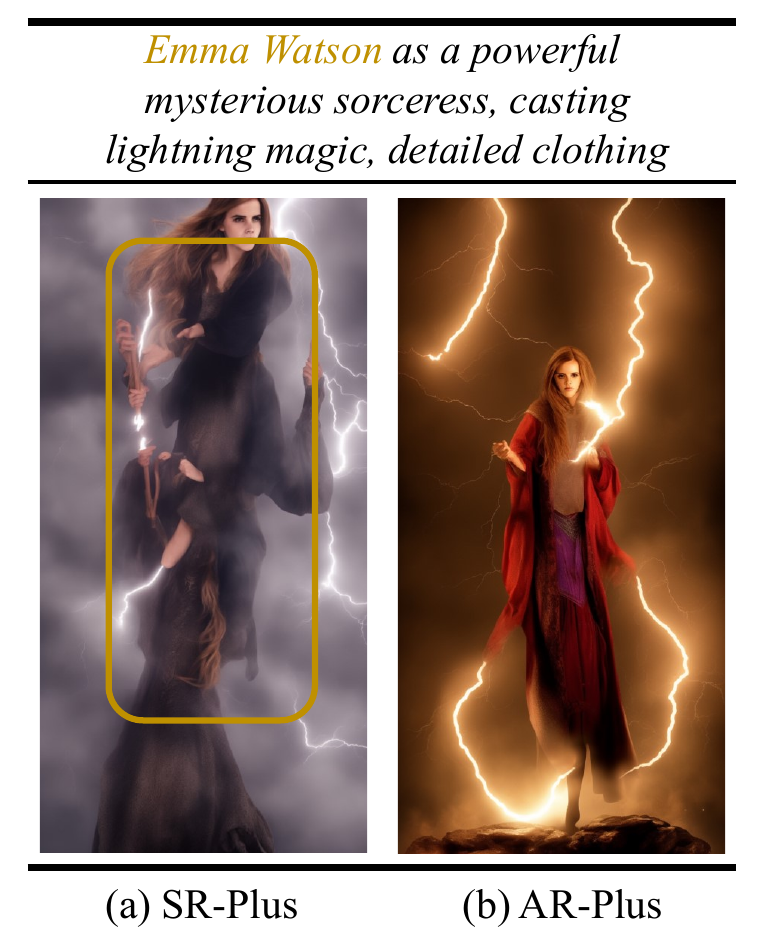}
    \vspace{-8.0mm}
    \caption{\textbf{Visual comparison:} SR-Plus (original SD 2.1) vs. AR-Plus (fine-tuned SD 2.1 with multi-aspect ratio training). The yellow box means poor composition.\vspace{-3.0mm}}
    \label{fig:supple1}
\end{figure}

\paragraph{Performance on the generation of images with different resolutions.}
Fig.~\ref{fig:resolutions} highlights the comparative efficiency of our proposed Any Size Diffusion (ASD) method relative to the baseline SR-Plus model, operating under a 32G GPU memory constraint. Specifically, in the context of generating images with dimensions of $1024^2$ and $2048^2$, our ASD algorithm consistently exhibits more efficient GPU memory usage than the SR-Plus model. For instance, when generating images of size $1024^2$, ASD consumes 13.44G of memory compared to SR-Plus's 17.45G; for $2048^2$ images, the consumption is 27.73G for ASD versus 31.69G for SR-Plus. Importantly, while the SR-Plus model is constrained to a maximum resolution of $2048^2$—beyond which it exceeds available GPU memory—our ASD method is developed to accommodate image resolutions of up to $18432^2$. This represents a significant 9$\times$ increase over the SR-Plus model's maximum capacity.

\begin{figure*}
    \centering
    \includegraphics[width=1.0\linewidth]{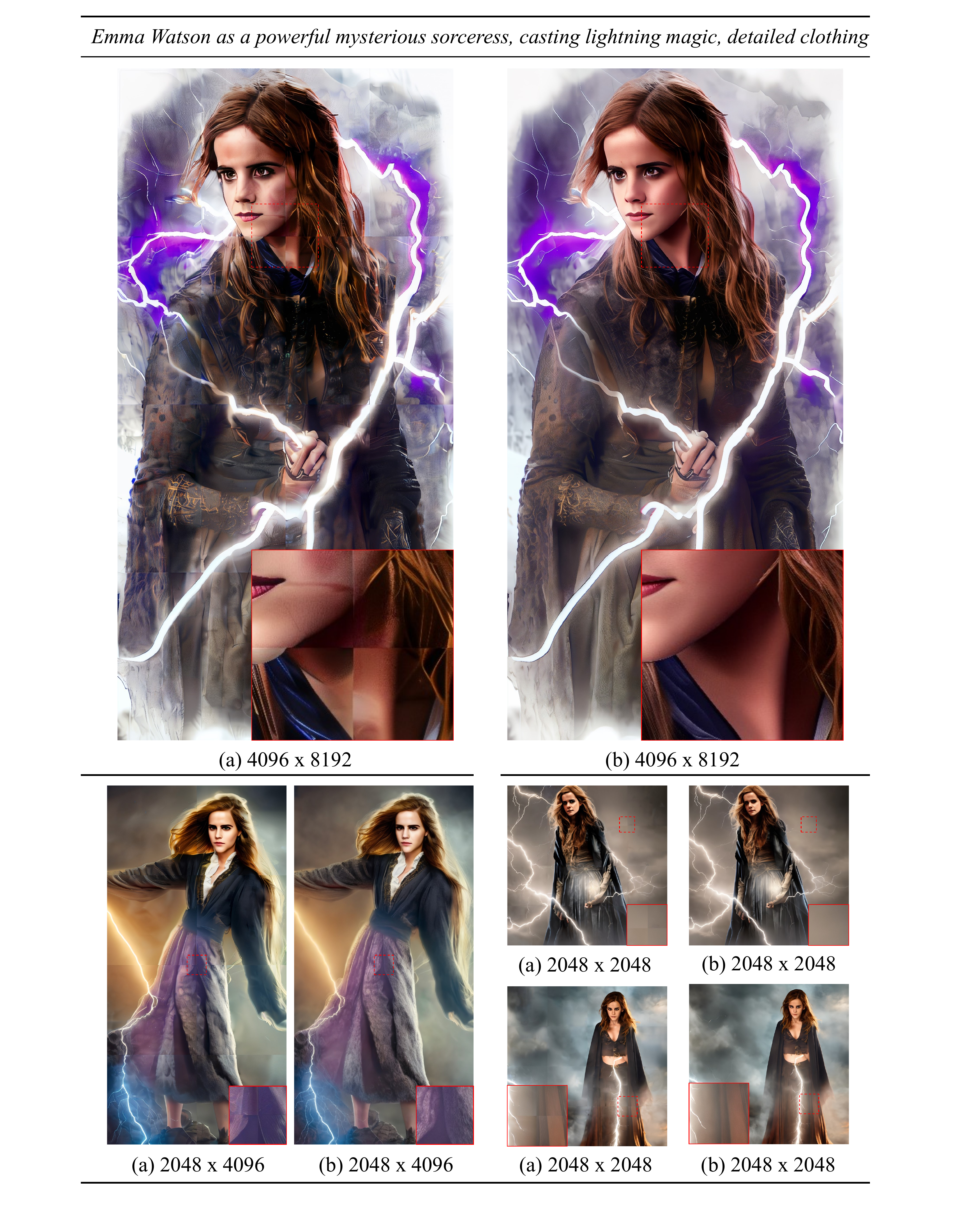}
    \vspace{-3.5em}
    \caption{\textbf{Comparative visualization of tiled diffusion techniques:} (a) AR-Tile without overlaps vs. (b) our ASD with implicit overlaps. Our ASD method enables the generation of 4K and 8K images while effectively avoiding seam issues.
\vspace{-1.0em}}
    \label{fig:supple2}
\end{figure*}

\paragraph{Achievement of 4K and 8K image-resolutions.} 
Fig.~\ref{fig:supple1} demonstrates the enhanced high-resolution image generation capabilities of our ASD method by contrasting it with SR-Plus, the original SD 2.1 model~\cite{Rombach_2022_CVPR}. SR-Plus degrades in composition for resolutions exceeding $512 \times 512$ pixels. In comparison, AR-Plus, developed through multi-aspect ratio training of SD 2.1, addresses this degradation but is bounded to a $2048 \times 2048$ pixel output under a 32GB GPU constraint. 
A non-overlapping tiled algorithm improves this limitation but introduces seaming artifacts, as shown in Fig.~\ref{fig:supple2}(a). Our solution, implementing implicit overlap in tiled sampling, resolves this seaming issue, with results depicted in Fig.~\ref{fig:supple2}(b). Thus, 
our ASD method effectively generates high-quality 4K and 8K images.


\bibliography{aaai24}

\bigskip

\end{document}